\documentclass[10pt,journal,compsoc]{IEEEtran}

\ifCLASSOPTIONcompsoc
  \usepackage[nocompress]{cite}
  \usepackage{multirow}
  \usepackage{array}
  \usepackage{graphicx,float}
  \usepackage{amsmath,amssymb,amsfonts}
  \usepackage{commath}
  \usepackage{afterpage}
  \usepackage[ruled,vlined]{algorithm2e}
  \usepackage{booktabs}
  \usepackage{caption}
  \usepackage{tablefootnote}
  \usepackage[dvipsnames]{xcolor}
  \graphicspath{{figure/}}
\else
  \usepackage{cite}
\fi
\ifCLASSINFOpdf
\else
\fi
\begin{document}
%
\title{Progressive Tandem Learning for Pattern Recognition with Deep Spiking Neural Networks}
%
%
%
%

\author{Jibin~Wu,
	Chenglin~Xu,
	Daquan~Zhou,
	Haizhou~Li,
	and~Kay~Chen~Tan
\IEEEcompsocitemizethanks{
	\IEEEcompsocthanksitem J.~Wu, D.~Zhou and H. Li are with the Department of Electrical and Computer Engineering, National University of Singapore, (e-mail: jibin.wu@u.nus.edu, daquan.zhou@u.nus.edu, haizhou.li@nus.edu.sg).
	\IEEEcompsocthanksitem C.~Xu is with the School of Computer Science and Engineering and Temasek Laboratories @ NTU, Nanyang Technological University, Singapore (e-mail:xuchenglin@ntu.edu.sg)
	\IEEEcompsocthanksitem K.~C.~Tan is with the Department of Computer Science, City University of Hong Kong, Hong Kong, (e-mail: kaytan@cityu.edu.hk).}
}

\SetKwInput{KwInput}{Input}                
\SetKwInput{KwOutput}{Output}              

\IEEEtitleabstractindextext{%
\begin{abstract}
Spiking neural networks (SNNs) have shown clear advantages over traditional artificial neural networks (ANNs) for low latency and high computational efficiency, due to their event-driven nature and sparse communication. However, the training of deep SNNs is not straightforward. In this paper, we propose a novel ANN-to-SNN conversion and layer-wise learning framework for rapid and efficient pattern recognition, which is referred to as progressive tandem learning of deep SNNs. By studying the equivalence between ANNs and SNNs in the discrete representation space, a primitive network conversion method is introduced that takes full advantage of spike count to approximate the activation value of analog neurons. To compensate for  the approximation errors arising from the primitive network conversion, we further introduce a layer-wise learning method with an adaptive training scheduler to fine-tune the network weights. The progressive tandem learning framework also allows hardware constraints, such as limited weight precision and fan-in connections, to be progressively imposed during training. The SNNs thus trained have demonstrated remarkable classification and regression capabilities on large-scale object recognition, image reconstruction, and speech separation tasks, while requiring at least an order of magnitude reduced inference time and synaptic operations than other state-of-the-art SNN implementations. It, therefore, opens up a myriad of opportunities for pervasive mobile and embedded devices with a limited power budget.
\end{abstract}

\begin{IEEEkeywords}
Deep Spiking Neural Network, ANN-to-SNN Conversion, Spike-based Learning, Large-scale Object Recognition, Speech Separation, Efficient Neuromorphic Inference
\end{IEEEkeywords}}

\maketitle

\IEEEdisplaynontitleabstractindextext

%
\IEEEpeerreviewmaketitle

\IEEEraisesectionheading{\section{Introduction}\label{sec:introduction}}
\IEEEPARstart{H}{uman} brains, after evolving for many hundreds of millions of years, are incredibly efficient and capable of performing complex pattern recognition tasks. In recent years, the deep artificial neural networks (ANNs) that are inspired by the hierarchically organized cortical networks have become the dominant approach for many pattern recognition tasks and achieved remarkable successes in a wide spectrum of application domains, instances include speech processing \cite{xiong2017toward,van2016wavenet}, computer vision \cite{krizhevsky2012imagenet,he2016deep}, language understanding \cite{hirschberg2015advances} and robotics \cite{silver2017mastering}. The deep ANNs, however, are notoriously expensive to operate both in terms of computational cost and memory usage. Therefore, they are prohibited from large-scale deployments in pervasive mobile and Internet-of-Things (IoT) devices.

In contrast, the adult's brains only consume about 20 watts to perform complex perceptual and cognitive tasks that are only equivalent to the power consumption of a dim light bulb \cite{laughlin2003communication}. While many efforts are devoted to improving the memory and computational efficiency of deep ANNs, for example, network compression \cite{han2015deep}, network quantization \cite{courbariaux2016binarized} and knowledge distillation \cite{hinton2015distilling}, it is more interesting to exploit the efficient computation paradigm inherent to the biological neural systems that are fundamentally different from and potentially integratable with the aforementioned strategies. 

The spiking neural networks (SNNs) are initially introduced to study the functioning and organizing mechanisms of biological brains. Recent studies have shown that deep ANNs also benefit from biologically realistic implementation, such as event-driven computation and sparse communication~\cite{pfeiffer2018deep}, for computational efficiency. Neuromorphic computing (NC), as an emerging non-von Neumann computing paradigm, aims to mimic the biological neural systems with SNNs in silicon \cite{roy2019towards}. The novel neuromorphic computing architectures, including Tianjic \cite{pei2019towards}, TrueNorth \cite{merolla2014million}, and Loihi \cite{davies2018loihi}, have shown compelling throughput and energy-efficiency in pattern recognition tasks, crediting to their inherent event-driven computation and fine-grained parallelism of the computing units. Moreover, the co-located memory and computation can effectively mitigate the problem of low bandwidth between the computing units and memory (i.e., von Neumann bottleneck) in data-driven pattern recognition tasks. 

It remains a challenge to train large-scale spiking neural networks that can be deployed onto these NC chips for real-world pattern recognition tasks. Due to the discrete and hence non-differentiable nature of spiking neuronal function, the powerful back-propagation (BP) algorithm that is widely used for deep ANN training is not directly applicable to the SNN. 

Recent studies suggest that the dynamical system formed by spiking neurons can be formulated as a recurrent ANN \cite{neftci2019surrogate}, whereby the subthreshold membrane potential dynamics of these leaky integrators (i.e., spiking neurons) can be effectively modeled. In addition, the discontinuity of spike generation function can be circumvented with surrogate gradients that provide an unbiased estimation of the true gradients \cite{wu2018direct,spatiotemporal,shrestha2018slayer,zenke2018superspike,STCA,bellec2018long}. In this way, the conventional error back-propagation through time algorithm (BPTT) can be applied to optimize the SNN. However, it is both computation- and memory-inefficient to optimize the SNN using the BPTT algorithm since spike trains are typically very sparse in both time and space. Therefore, the scalability of the technique remains to be improved, for instance, the size of SNNs is GPU memory bounded as demonstrated in a gesture classification task \cite{shrestha2018slayer}. Furthermore, the vanishing and exploding gradient problem \cite{hochreiter1998vanishing} of the BPTT algorithm adversely affects the learning in face of input spike trains of long temporal duration or low firing rate. 

To address the aforementioned issues in surrogate gradient learning, a novel tandem learning framework \cite{wu2019hybrid} has been proposed. This learning framework consists of an ANN and an SNN coupled through weight sharing, where the SNN is used to derive the exact neural representation, while the ANN is designed to approximate the surrogate gradients at the spike-train level. The SNNs thus trained have demonstrated competitive classification and regression capabilities on a number of frame- and event-based benchmarks, with significantly reduced computational cost and memory usage. Despite the promising learning performance demonstrated by these spike-based learning methods, their applicability to deep SNNs with more than 10 hidden layers remains elusive. 

Following the idea of rate-coding, recent studies have shown that SNNs can be effectively constructed from ANNs by approximating the activation value of analog neurons with the firing rate of spiking neurons\cite{cao2015spiking,diehl2015fast,ethImageNet,sengupta2019going,kim2019spiking,hu2018spiking}. 
This approach not only simplifies the training procedures of aforementioned spike-based learning methods but also enable SNNs to achieve the best-reported results on a number of challenging tasks, including object recognition on the ImageNet-12 \cite{ethImageNet,sengupta2019going} dataset and object detection on the PASCAL VOC and MS COCO datasets \cite{kim2019spiking}. However, to reach a reliable firing rate approximation, it requires a notoriously large encoding time window with at least a few hundreds of time steps. Moreover, the total number of synaptic operations required to perform one classification usually increases with the size of the encoding time window, therefore, a large encoding time window will also adversely impact the computational efficiency. 
An ideal SNN model should not only perform pattern recognition tasks with high accuracy but also obtained the results rapidly with as few time steps as possible and efficiently with a small number of synaptic operations. In this work, we introduce a novel ANN-to-SNN conversion and learning framework to progressively convert a pre-trained ANN into an SNN for accurate, rapid, and efficient pattern recognition. 

To improve the inference speed and energy efficiency, we introduce a layer-wise threshold determination mechanism to make good use of the encoding time window of spiking neurons for information representation. To maintain a high pattern recognition accuracy, a layer-wise learning method with an adaptive training scheduler is further applied to fine-tune the network weights after each primitive layer conversion that compensates for the conversion errors. The proposed layer-wise conversion and learning framework also supports effective algorithm-hardware co-design by progressively imposing hardware constraints during training. To summarize, the main contributions of this work are in four aspects:
\begin{itemize}
\item \textbf{Rethinking ANN-to-SNN Conversion:} We introduce a new perspective to understand the neural discretization process of spiking neurons by comparing it to the activation quantization of analog neurons, which offers a new angle to understand and perform network conversion. By making efficient use of the spike count that upper bounded by the encoding time window size to represent the information of analog counterparts, the inference speed and computational cost can be significantly reduced over other conversion methods that grounded on a firing rate approximation.

\item \textbf{Progressive Tandem Learning Framework:} We propose a novel layer-wise ANN-to-SNN conversion and learning framework with an adaptive training scheduler to support effortless and efficient conversion, which allows fast, accurate, and efficient pattern recognition with deep SNNs. The proposed conversion framework also allows easy incorporation of hardware constraints into the training process, for instance, limited weight precision and fan-in connections, such that the optimal performance can be achieved when deploying onto the actual neuromorphic chips.

\item \textbf{Rethinking Spike-based Learning Methods:} We conduct a comprehensive study on the scalability of both the time-based surrogate gradient learning and the spike count-based tandem learning methods, revealing that the accumulated gradient approximation errors may impede the training convergence in deep SNNs.

\item \textbf{Solving Cocktail Party Problem with SNN:} To evaluate the proposed learning framework, we apply deep SNNs to separate high fidelity voices from a mixed multiple talker speech, which effectively mimics the perceptual and cognitive ability of the human brain. To the best of our knowledge, this is the first work that successfully applied deep SNNs to solve the challenging cocktail party problem.
\end{itemize}

The rest of the paper is organized as follows. In Section \ref{sec:related_works}, we first review the conventional ANN-to-SNN conversion methods and discuss the trade-off between accuracy and latency. In Section \ref{sec:ann_to_snn_conversion}, we compare the neuronal functions between the spiking neurons and analog neurons, and their discrete equivalents, that provide a new perspective to perform network conversion. With this, we propose to use the spike count of spiking neurons as the bridge between the spiking neurons and their analog counterparts for network conversion. In Section \ref{sec:layer_wise_training}, to minimize the conversion errors, we propose a novel layer-wise learning method with an adaptive training scheduler to fine-tune network weights. In Sections \ref{sec:experiment_classification} and \ref{sec:experiment_regression}, we validate the proposed network conversion and learning framework, that is referred to as  \textit{progressive tandem learning} (PTL), through a set of classification and regression tasks, including the large-scale image classification, time-domain speech separation and image reconstruction. Finally, we conclude the paper in Section \ref{sec:conclusion}.

\section{Related Work}
\label{sec:related_works} 
Recently, many ANN-to-SNN conversion methods are proposed. Nearly all of these methods follow the idea of rate-coding, which approximates the activation value of analog neurons with the firing rate of spiking neurons. In what follows, we will review the development of ANN-to-SNN conversion methods and highlight the issue of accuracy and latency trade-off in these methods.

The earliest attempt for ANN-to-SNN conversion was presented in \cite{6497055}, where Pérez-Carrasco et al. devised an approximation method for leaky integrate-and-fire (LIF) neurons using analog neurons. The pre-trained weights of analog neurons are rescaled by considering the leaky rate and other parameters of spiking neurons before copying into the SNN. This conversion method was proposed to handle event streams captured by the event-driven camera, whereby promising recognition results were demonstrated on the human silhouette orientation and poker card symbol recognition tasks. While this conversion method requires a large number of hyperparameters to be determined manually and the conversion process suffers from the quantization and other approximation errors.

There are recent studies on ANN-to-SNN conversion with applications to accurate object recognition and detection on the frame-based images.  Cao et al. \cite{cao2015spiking} proposed a conversion framework by using the rectified linear unit (ReLU) as the activation function for analog neurons and set the bias term to zero. The activation value of analog neurons can thus be well approximated by the firing rate of integrate-and-fire (IF) neurons. Furthermore, the max-pooling operation, which is hard to determine in the temporal domain for a rate-based SNN, is replaced with the average pooling. Diehl et al. \cite{diehl2015fast} further improved this conversion framework by analyzing the causes of performance degradation, which reveals the potential problems of over- and under-activation of spiking neurons. To address these problems, they proposed model- and data-based weight normalization schemes to rescale the SNN weights based on the maximum activation values of analog neurons. These normalization schemes prevent the over- and under-activation of spiking neurons and strike a good balance between the firing threshold and the model weights. As a result, near-lossless classification accuracies were reported on the MNIST dataset with fully-connected and convolutional spiking neural networks.

Rueckauer et al. \cite{ethImageNet} identified a quantization error caused by the reset-to-zero scheme of IF neurons, where surplus membrane potential over the firing threshold is discarded after firing. This quantization error tends to accumulate over layers and severely impact the classification accuracy of converted deep SNNs. To address this problem, they propose a reset-by-subtraction scheme to preserve the surplus membrane potential after each firing. Moreover, a modified data-based weight normalization scheme is introduced to improve the robustness against outliers, which significantly improves the firing rate of spiking neurons and hence the inference speed of SNN. For the first time, they had demonstrated competitive results to the ANN counterparts on the challenging ImageNet-12 object recognition task.

In the same line of research, Hu et al. \cite{hu2018spiking} provided a systematic approach to convert deep residual networks and propose an error compensation scheme to address the accumulated quantization errors. With these modifications, they achieved near-lossless conversion for spiking residual networks up to 110 layers. Kim et al. \cite{kim2019spiking} extended the conversion framework by applying the weight normalization channel-wise for convolutional neural networks and propose an effective strategy for converting analog neurons with both positive and negative activation values. The proposed channel-wise normalization scheme boosted the firing rate of neurons and hence improved the information transmission rate. Benefiting from these modifications, competitive results are demonstrated in the challenging objection detection task where the precise coordinate of bounding boxes is required to be predicted. Sengupta et al. \cite{sengupta2019going} further optimized the weight normalization scheme by taking into consideration the behavior of spiking neurons at the run time, which achieved the best reported result on the ImageNet-12 dataset. 

In these earlier studies, methods are proposed for the firing threshold determination or weight normalization so as to achieve a good firing rate approximation. Despite competitive results achieved by these conversion methods, the underlying firing-rate assumption has led to an inherent trade-off between accuracy and latency, which requires a few hundred to thousands of time steps to reach a stable firing rate. Rueckauer et al. \cite{ethImageNet} provided a theoretical analysis of this issue by analyzing the firing rate deviation of these ANN-to-SNN conversion methods. By assuming a constant input current to spiking neurons at the first layer, the actual firing rate of the first (Eq. \ref{eq:firing_rate_error_first}) and subsequent layers (Eq. \ref{eq:firing_rate_error_top}) can be summarised as follows
\begin{equation}
{\rm{r}}_i^1(t) = a_i^1{r_{\max }} - \frac{{V_i^1(t)}}{{t\vartheta }}
\label{eq:firing_rate_error_first}
\end{equation}

\begin{equation}
{\rm{r}}_i^l(t) = \sum\limits_j^{{M^{l - 1}}} {w_{ij}^l} {\rm{r}}_j^{l - 1}(t) + b_i^l{{\rm{r}}_{\max }} - \frac{{V_i^l(t)}}{{t\vartheta }}
\label{eq:firing_rate_error_top}
\end{equation}
where ${\rm{r}}_i^l(t)$ denotes the firing rate of neuron $i$ at layer $l$ and $r_{\max }$ denotes the maximum firing rate that is determined by the time step size. $a_i^1$ is the activation value of analog neuron $i$ at the first layer and
$V_i^1(t)$ is the membrane potential of the corresponding spiking neuron. $M^{l - 1}$ is the total number of neurons in layer $l-1$ and $b_i^l$ is the bias term of analog neuron $i$ at layer $l$. Ideally, the firing rate of spiking neurons should be proportional to the activation value of their analog counterparts as per the first term of Eq. \ref{eq:firing_rate_error_first}. While the surplus membrane potential that has not been discharged by the end of simulation will cause an approximation error as shown by the second term of Eq. \ref{eq:firing_rate_error_first}, which can be counteracted with a large firing threshold or a large encoding time window. Since increasing the firing threshold will inevitably prolong the evidence accumulation time, a proper firing threshold that can prevent spiking neurons from either under- or over-activating is usually preferred and the encoding time window is extended to minimize such a firing rate approximation error. 

Besides, this approximation error accumulates gradually while propagating over layers as shown in Eq. \ref{eq:firing_rate_error_top}, thereby further extension of encoding time window is required to compensate. As such, a few thousands of time steps are typically required to achieve a competitive accuracy with deep SNNs of more than 10 layers \cite{kim2019spiking, sengupta2019going}. From these formulations, it is clear that to approximate the continuous input-output representation of ANNs with the firing rate of spiking neurons will inevitably lead to the accuracy and latency trade-off. To overcome this issue, as will be introduced in the following sections, we propose a novel conversion method that is grounded on the discrete neural representation, whereby the spike count that upper bounded by the encoding time window is taken to approximate the discrete input-output representation of ANNs. By determining the firing threshold of spiking neurons to make efficient use of the encoding time window, the rapid and efficient pattern recognition can be achieved with SNNs. To counteract the conversion errors and hence ensure high accuracies in pattern recognition tasks, a layer-wise learning method is further proposed to fine-tune the network weights.

\section{Rethinking ANN-to-SNN Conversion}
\label{sec:ann_to_snn_conversion}
Over the years, many spiking neuron models are developed to describe the rich dynamical behavior of biological neurons. Most of them, however, are too complex for real-world pattern recognition tasks. As discussed in Section \ref{sec:related_works}, for computational simplicity and ease of conversion, 
the integrate-and-fire (IF) neuron model is commonly used in ANN-to-SNN conversion works~\cite{diehl2015fast,ethImageNet,sengupta2019going}. Although this simplified spiking neuron model does not emulate the rich sub-threshold dynamics of biological neurons, it preserves attractive properties of discrete and sparse communication, therefore, allows for efficient hardware implementation. In this section, we reinvestigate the approximation of input-output representation between a ReLU analog neuron and an integrate-and-fire spiking neuron.

\subsection{Spiking Neuron vs Analog Neuron}
Let us consider a discrete-time simulation of spiking neurons with an encoding time window of $N_{s}$ that determines the inference speed of an SNN. At each time step $t$, the incoming spikes to the neuron $i$ at layer $l$ are transduced into synaptic current $z_i^l[t]$ according to
\begin{equation}
z_i^l[t] = \sum\nolimits_j {w_{ij}^{l - 1}s_j^{l - 1}} [t] + b_i^l\\
\label{eq1}
\end{equation}
where $s_j^{l - 1}[t]$ indicates the occurrence of an input spike at time step $t$, and $w_{ij}^{l - 1}$ is the synaptic weight between the pre-synaptic neuron $j$ and the post-synaptic neuron $i$ at layer $l$. $b_i^l$ can be interpreted as a constant injecting current. 

The synaptic current $z_i^l[t]$ is further integrated into the membrane potential $V_i^l[t]$ as per Eq. \ref{eq2}. Without loss of generality, a unitary membrane resistance is assumed in this work. The membrane potential is reset by subtracting the firing threshold after each firing as described by the last term of Eq. \ref{eq2}. 
\begin{equation}
V_i^l[t] = V_i^l[t - 1] + z_i^l[t] - \vartheta^l s_i^l[t - 1]
\label{eq2}
\end{equation}

An output spike is generated whenever the $V_i^l[t]$ rises above the firing threshold $\vartheta^l$ (determined layer-wise) as follows
\begin{equation}
s_i^l[t] = \Theta (V_i^l[t] - \vartheta^l) \;\; with \;\; \Theta (x) = \left\{\begin{array}{l}1,\;\;\; if\;x \ge 0\\
0,\;\;\;otherwise\;
\end{array} \right.
\label{eq3}
\end{equation}

The spike train ${s_i^l}$ and spike count $c_i^{l}$ for a time window of $N_s$ can thus be determined and represented as follows
\begin{equation}
\begin{split}
{s_i^l} &= \{{s_i ^l}[1],...,{s_i ^l}[{N_{s}}]\} \\
c_i^{l} &= \sum\nolimits_{t=1}^{N_{s}} {s_i^{l}[t]} 
\label{eq4}
\end{split}
\end{equation}

For non-spiking analog neurons, let us describe the neuronal function of neuron $i$ at layer $l$ as
\begin{equation}
a_i^l = f({\sum\nolimits_j {w_{ij}^{l - 1}} x_j^{l - 1} + b_i^l})
\label{analog_neuron}
\end{equation}
which has $w_{ij}^{l - 1}$ and $b_i^l$ as the weight and bias. $x_j^{l - 1}$ and $a_i^l$ denote the input and output of the analog neuron. $f(\cdot)$ denotes the activation function, which we use the ReLU in this work. For ANN-to-SNN conversion, an ANN with the ReLU analog neurons is first trained, that is called pre-training, before the conversion.

\subsection{Neural Discretization vs Activation Quantization}
In the conventional ANN-to-SNN conversion studies, the firing rate of spiking neurons is usually taken to approximate the continuous input-output representation of the pre-trained ANN. As discussed in Section \ref{sec:related_works}, a spiking neuron takes a notoriously long time window to reliably approximate a continuous value.  Recent studies, however, suggest such a continuous neural representation may not be necessary for ANNs \cite{van2017neural}. In fact, there could be little impact on the network performance when the activation value of analog neurons are properly quantized to a low-precision discrete representation \cite{jacob2018quantization,severa2019training}, which is known as activation quantization. 

In ANNs, the activation quantization refers to the mapping of a floating-point activation value $a_{i}^{l,f}$ to a quantized value $a_{i}^{l,q}$. With a ReLU activaiton function, the activation quantization can be formulated as follows
\begin{equation}
\begin{split}
{\hat{a}_{i}^{l,f}} &= \min (\max ({a_{i}^{l,f}},0),a_u^{l})\\
\varphi^l &= \frac{a_u^{l}}{N_q}\\
{a_{i}^{l,q}} &= round\left( {\frac{{\hat{a}_{i}^{l,f}}}{\varphi^l}} \right) \cdot \varphi^l 
\end{split}
\label{quantize}
\end{equation}
where $a_u^{l}$ refers to the upper bound of the quantization range at layer $l$, whose values are usually determined from the training data. $N_q$ is the total number of quantization levels and $\varphi^l$ is the quantization scale for layer $l$. With such a discrete neural representation, the computation and storage overheads during training and inference of ANNs can be significantly reduced. The success of activation quantization can be explained by the fact that there is a high level of redundancy in the continuous neural representation.  

In SNNs, the information is inherently discretized into spike trains according to the neuronal dynamics of spiking neurons, which is referred to as the neural discretization hereafter. It worth noting that the size of the encoding time window determines the discrete representation space for SNNs. The activation quantization of ANNs leads to a reduction in data storage, which takes place in the spatial domain. By mapping the discrete neural representation of a good performing ANN to an SNN, it is expected that we translate the reduction of the data storage into the reduction of the encoding time window size, thus allowing rapid and efficient pattern recognition with SNNs.

The analog neurons respond to the input stimuli instantly, while spiking neurons respond to the input spike trains through a temporal process within a time window. In order to establish a correspondence between the activation quantization of analog neurons and the neural discretization of spiking neurons, we simplify the neural discretization process by assuming the preceding layer's spike trains and the constant injecting current are integrated and discharged instantly. 
The overall contributions from the preceding layer's spike trains and constant injecting current can be summarized by the free aggregate membrane potential (no firing) \cite{wu2019hybrid} defined as 
\begin{equation}
V_i^{l} = \sum\nolimits_j {w_{ij}^{l - 1}c_j^{l - 1}} + b_i^l{N_s} \\
\label{eq5}
\end{equation}
By considering $b_i^l{N_s}$ as the bias term and $c_j^{l - 1}$ as the input to analog neurons that defined in Eq. \ref{analog_neuron}, $V_i^{l}$ is exactly the same as the pre-activation quantity of non-spiking analog neurons. By considering the spike count of spiking neurons as the information carrier, the simplification of neural discretization provides the basis for mapping the discrete inputs of an analog neuron to the discrete spike count inputs of a spiking neuron.

Note that an IF neuron responds to the input spike trains by firing zero or a positive number of output spikes. It performs a non-linear transformation similar to that of the ReLU activation function of an analog neuron. As defined in Eq. \ref{quantize}, the activation quantization discretizes the positive activation value of ReLU neurons, by a fixed quantization scale $\varphi^l$, into an integer. Similarly, the neural discretization of an IF neuron discretizes the positive-valued $V_i^{l}$ by a fixed discretization scale, that is the firing threshold $\vartheta^l$, into a discrete spike count, that can be formulated as follows 
\begin{equation}
\begin{split}
{\hat{V}_i^{l}} &= \min (\max ({V_i^{l}},0),V_u^{l})\\
\vartheta^l &= \frac{V_u^{l}}{{{N_s}}}\\
V_{i}^{l,q} &= \underbrace {round\left( {\frac{{{\hat{V}_i^{l}}}}{\vartheta^l}} \right)}_{ \approx c_i^l} \cdot \vartheta^l
\end{split}
\label{quantize_snn}
\end{equation}
where $V_u^{l}$ refers to the free aggregated membrane potential upper bound of layer $l$. The Eqs. \ref{quantize} and \ref{quantize_snn} establish a correspondence between the activation quantization of a ReLU neuron and the discrete neural representation of an IF neuron, thus provides the basis for mapping the discrete output of an analog neuron to the spike count output of a spiking neuron. It is worth noting that the quantization scale $\varphi^l$ is usually stored independently for ANNs and multiplied to the fixed point number during operations. However, the discretization scale $\vartheta^l$ is only stored at the spiking neuron and does not propagate together with output spike trains to the next layer. This issue can be easily counteracted by multiplying $\vartheta^l$ to the weights of the next layer.

\begin{figure}[htb]
	\centering
	\centerline
	{\includegraphics[width = 9 cm]{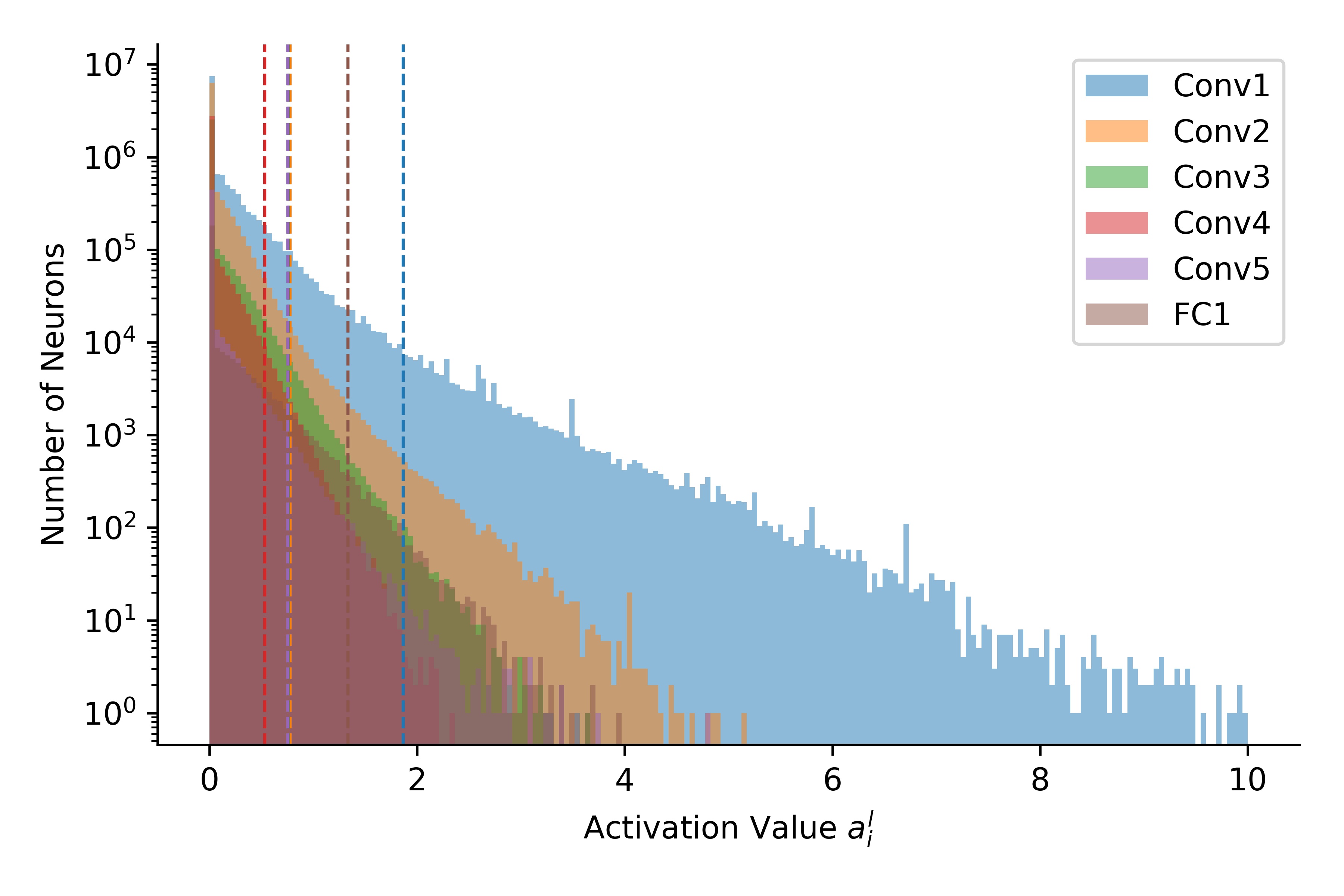}}
	\caption{Distribution of the activation value $a_i^{l}$ of ReLU neurons in the pre-trained ANN layers. Here, the horizontal axis represents the activation values, while the vertical axis represents the number of neurons in a log scale. The majority of neurons output low activation values and the number of neurons decreases rapidly as the activation value increases. The dotted lines mark the 99th percentile of the number of neurons in each layer.}
	\label{act_dist}
\end{figure}

With the simplification of neural discretization, we show that the discrete input-output representation of analog neurons can be well approximated with spiking neurons. Following this formulation, an SNN can be constructed from the pre-trained ANN by directly copying its weights. The constant injecting current to spiking neurons can be determined by dividing the bias term of the corresponding analog neuron over $N_s$. According to Eq. \ref{quantize_snn}, the firing threshold $\vartheta^l$ of spiking neurons at layer $l$ can be determined by dividing the upper bound $V_u^{l}$ over $N_s$. From Eqs. \ref{analog_neuron} and \ref{eq5}, it clear that the upper bound $V_u^{l}$ is equivalent to and hence can be directly taken from the maximum activation value $a_u^{l}$ of the corresponding ANN layer.

However, two potential errors may arise from this formulation: a quantization error affected by the encoding time window size and a spike count approximation error arisen from the temporal structure of input spike trains that may affect the discharging of $V_i^{l}$. These conversion errors, however, can be effectively mitigated by the threshold normalization mechanism and the layer-wise training method that will be introduced in the following sections.

\subsection{Threshold LayerNorm}
To better represent the quantization range of analog neurons using spiking neurons that have a pre-defined encoding time window $N_s$, we introduce a novel threshold determination mechanism for spiking neurons. To properly define the quantization range of analog neurons in a layer, we need to determine the activation value upper bound $a_u^{l}$. 

\begin{figure*}[htb]	
	\centering
	\centerline
	{\includegraphics[width = 18 cm]{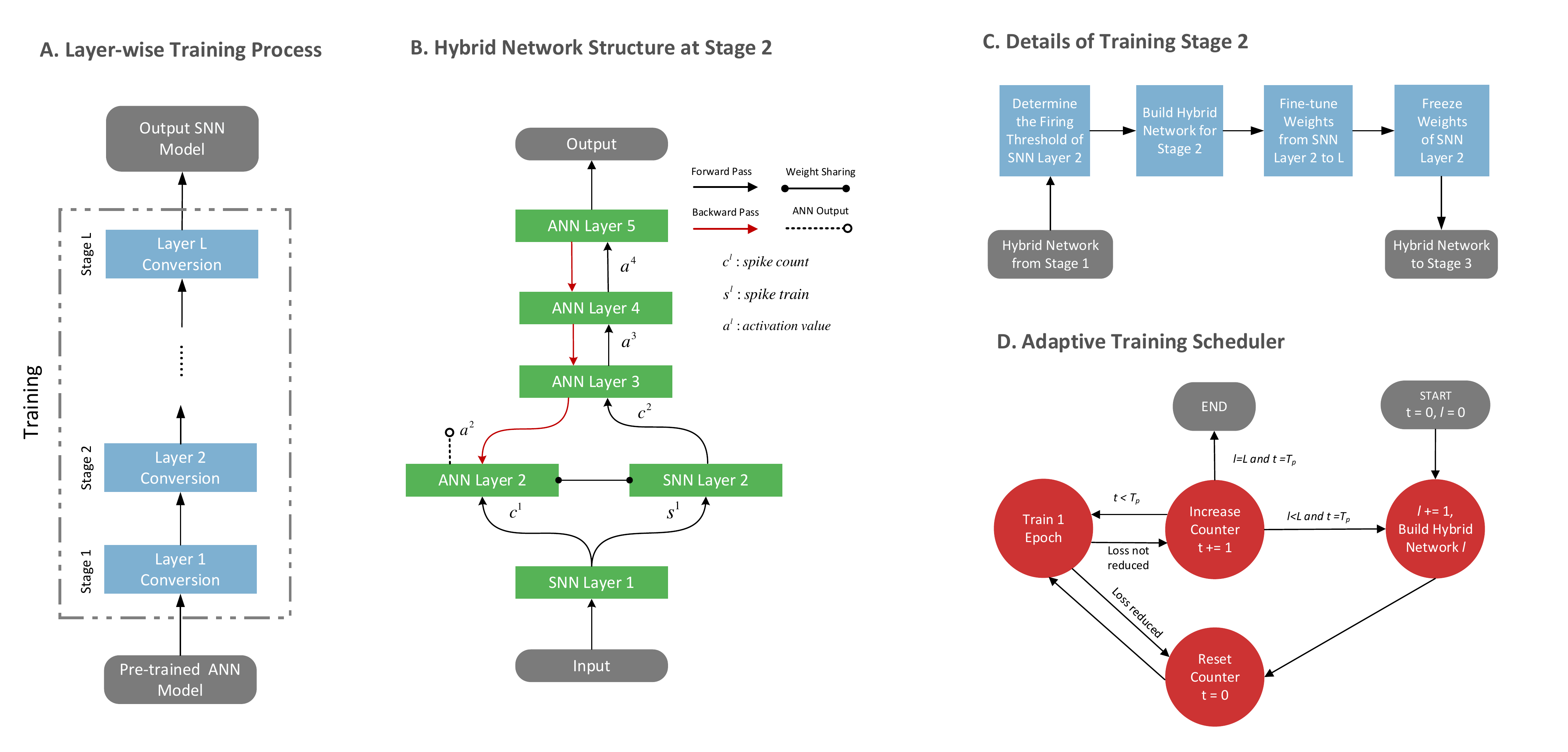}}
	\caption{Illustration of the proposed PTL framework. (A) The whole training process is organized into separate stages. (B) Details of the hybrid network at the training stage 2. (C) Details of the training processes at stage 2. (D) Illustration of the adaptive training scheduler.}
	\label{learning_rule}
\end{figure*}

As shown in Fig. \ref{act_dist} and also highlighted in \cite{ethImageNet}, the $a_u^{l}$ tends to be biased by the outlier samples, for instance, the $a_u^{1}$ of Conv1 layer is five times higher than the 99th percentile (highlighted as the blue dotted line). To make efficient use of the available discrete representation space and reduce the quantization errors, we propose to use the 99th or 99.9th percentile of all $a_i^{l}$ in a layer, determined from the training data, as the upper bound $a_u^{l}$ such that the key information can be well-preserved. Given the equivalence of $a_u^{l}$ and $V_u^{l}$ established in the earlier section, the firing threshold  $\vartheta^l$ of spiking neurons at layer $l$ can hence be determined by dividing the value of $a_u^{l}$ over $N_s$. In practice, we observe these percentiles remain relatively stable across data batches with a sufficiently large batch size (e.g., 128 or 256). Therefore, the $a_u^{l}$ can be effectively derived from a random training batch. 

To further improve the numerical resolution for convolutional neural networks, the firing threshold can be determined independently for each channel similar to that proposed in \cite{kim2019spiking}. While we did not notice significant improvements in the classification or regression performance in our experiments, probably due to the layer-wise learning method that we have applied counteracts the performance drop. 

\subsection{Neural Coding}
\label{subsec:neural_code}
A suitable neural encoding scheme is required to convert the static input feature tensors or images into spike trains for neural processing in SNNs. It was found that a direct discretization of the inputs introduces significant distortions to the underlying information. While discretizing the feature tensors derived from the first network layer can effectively preserve the information by leveraging the redundancies in the high-dimensional feature representation \cite{anderson2017high}. Following this approach, we interpret the activation value $a_i^l$ of analog neurons as the input current to the corresponding spiking neurons and add it to Eq. \ref{eq2} at the first time step. The spike trains are generated by distributing this quantity over consecutive time steps according to the dynamic of IF neurons; the spiking output then starts from the first hidden layer. This neural encoding scheme effectively discretizes the feature tensor and represent it as spike counts. 

The neural decoding determines the output class from the synaptic activity of spiking neurons. Instead of using the discrete spike counts, we suggest using the free aggregate membrane potential of neurons in the final SNN layer to determine the output class, which provides a much smoother learning curve over the discrete spike count due to the continuous error gradients derived at the output layer \cite{wu2019hybrid}. Moreover, this continuous quantity can also be directly considered as the outputs in regression tasks, such as image reconstruction and speech separation that will be presented later in this paper.

\section{Progressive Tandem Learning}
\label{sec:layer_wise_training}
The primitive ANN-to-SNN conversion method introduced in the earlier section provides a more efficient way to approximate the input-output representation of ANNs. However, the conversion process inherently introduces quantization and spike count approximation errors as discussed in Section 3.2. Such errors tend to accumulate over layers and cause significant performance degradation especially with a small $N_s$. This therefore calls for a training scheme to fine-tune the network weights after the primitive conversion, so as to compensate for these conversion errors. 

There have been spike-based learning schemes, such as time-based surrogate gradient learning \cite{neftci2019surrogate} and spike count-based tandem learning methods \cite{wu2019hybrid}, for SNN training in an end-to-end manner. However, they don't work the best for the required fine-tuning task. For example, the surrogate gradients approximated from these methods tend to be noisy for an extremely short encoding time window, that we would like to have. As will be seen in Section \ref{sec:spike_based_learning}, gradient approximation errors accumulate over layers with these end-to-end learning methods, that significantly degrade the learning performance for an SNN of over 10 layers.

To address this issue, we propose a layer-wise learning method, whereby ANN layers are converted into SNN layers one layer at a time. We define the conversion and weight fine-tuning of one SNN layer as one stage. Therefore, for an ANN network of $L$ layers, as shown in Fig. \ref{learning_rule}(A), it takes $L$ stages to complete the entire conversion and fine-tuning process.

The details of each training stage are illustrated in Fig. \ref{learning_rule}(C). All spiking neurons in the same SNN layer share the same firing threshold, which is first determined according to the proposed Threshold LayerNorm mechanism. Besides, the constant injecting current to spiking neurons is determined by dividing the corresponding bias term of analog neurons over $N_s$. Following the tandem learning approach \cite{wu2019hybrid}, a hybrid network is further constructed by coupling the converted SNN layer to the pre-trained ANN layer through weight sharing, thereafter the ANN layer becomes an auxiliary structure to facilitate the fine-tuning of the converted SNN layer. At each training stage, the PTL scheme follows the tandem learning idea except that 1) we fix the weights of the SNN layers in the previous stages; 2) we update only one SNN layer together with all ANN layers.

\subsection{Tandem Learning}
As shown in Fig. \ref{learning_rule}(B), the spike trains, derived from the preceding SNN layer, and their equivalent spike counts are forward propagated to the coupled layer. In the coupled layer, the spiking neurons take spike trains as input and generate spike counts as output, while the analog neurons take spike counts as input and generate an output quantity that approximates the spike count of the coupled spiking neurons. To allow for weight sharing between the ANN and the SNN layers, we take the spike counts as the bridge. To this end, let us express the non-linear transformation of a spiking neuron as  
\begin{equation}
c_i^l = g({s^{l - 1}};w_i^{l - 1},b_i^l,\vartheta^l)
\end{equation}
where $g(\cdot)$ denotes the effective transformation performed by spiking neurons. Given the state-dependent nature of spike generation, it is not feasible to directly determine an analytical expression from $s^{l - 1}$ to $c_i^l$. Here, we simplify the spike generation process by assuming the resulting synaptic currents from $s^{l - 1}$ are evenly distributed over time. We thus obtain the interspike interval of the output spike train as
\begin{equation}
\Delta_i^l = \rho \left( {\frac{\vartheta^l{N_s} }{{(\sum\limits_j {w_{ij}^{l - 1}c_j^{l - 1} + b_i^l{N_s})} }}} \right)
\end{equation}
where $\rho(\cdot)$ denotes the ReLU non-linearity. The equivalent output spike count can be further determined as
\begin{equation}
c_i^l = \frac{{{N_s}}}{{\Delta_i^l}} = \frac{1}{\vartheta^l } \,\ \rho \left( {\sum\limits_j {w_{ij}^{l - 1}c_j^{l - 1} + b_i^l {N_s}} } \right)
\label{eq:spike_count}
\end{equation}

In practice, to reuse the original ANN layer for the fine-tuning purpose, we absorb the scaling factor $1/\vartheta^l$ into the learning rate. This configuration allows spike-train level error gradients to be effectively approximated from the ANN layer. It was shown that the ANN-SNN tandem learning method works more efficiently for rate-coded networks than other spike-based learning methods that update the weights for each time step \cite{wu2019hybrid}. 

In this paper, the tandem learning rule allows the spiking synaptic filters to be fine-tuned after the primitive conversion, which offers a good initialization for discrete neural representation. Along with the weights fine-tuning of subsequent ANN layers, the conversion errors can be effectively mitigated. Different from the end-to-end tandem learning framework introduced in \cite{wu2019hybrid}, the tandem learning here is performed one layer at a time to prevent the gradient approximation error from accumulating across layers. The weights of the SNN layer are frozen after each training stage.

\subsection{Scheduling of Progressive Tandem Learning}
The PTL framework requires a schedule to be determined for each training stage. Inspired from \cite{severa2019training}, we propose an adaptive training scheduler to automate the PTL process. As shown in Fig. \ref{learning_rule}(D), at the end of each training epoch we update the patience counter $t$ based on the current validation loss and the best validation loss at the current training stage. The patience counter is reset to zero when the current validation loss improves, otherwise, the patience counter is increased by one. Once the patience counter reaches the pre-defined patience period $T_p$, the hybrid network parameters with the best validation loss are re-loaded to the network (i.e., the best model at the current training stage) before the weights of the trained SNN layer are frozen. The training process terminates after the last ANN layer is replaced by the SNN layer. The pseudo codes of the proposed layer-wise ANN-to-SNN conversion framework are presented in Algorithm \ref{algo}.

\begin{algorithm}
	\small
	\DontPrintSemicolon
	\KwInput{input sample $X_{in}$, target label $Y$, pre-trained ANN $net_a$, encoding time window size $N_s$, patience period $T_p$, number of network layers $L$}
	\KwOutput{converted SNN}
	\vspace{4mm}
	\tcp*[l]{\scriptsize Network Initialization} 
	$net$ = $net_a$ \\
	\For{layer $l$ = 1 to $L$}
	{
		\tcp*[l]{\scriptsize Initialize the Training Scheduler} 
		$t$ = 1 \\
		$loss\_best = \infty$ \\
		\tcp*[l]{\scriptsize Determine the Firing Threshold of Layer $l$} 
		$\vartheta^l = Threshold\_LayerNorm(net, N_s, X_{in})$ \\
		\tcp*[l]{\scriptsize Build Hybrid Network for Training Stage $l$} 
		$net = Build\_Hybrid\_Network(net, \vartheta^l, l)$ \\
		\While{ $t < Tp$}{
			\tcp*[l]{\scriptsize Layer-wise Training for 1 Epoch on the Hybrid Network} 
			$[net, val\_loss] = Layer\_Wise\_Training(net, N_s, X_{in}, Y) $ 
			\tcp*[l]{\scriptsize Update the Training Scheduler} 
			$[t, loss\_best] = Update\_Training\_Scheduler(val\_loss, loss\_best) $ 
		}
		\tcp*[l]{\scriptsize Freeze the Weights of SNN Layer $l$} 
		$net = Freeze\_Layer(net, l) $ 
	} 
	\caption{Pseudo Codes of the Progressive Tandem Learning Framework}
	\label{algo}
\end{algorithm}

\subsection{Optimizing for Other Hardware Constraints}  
The PTL framework also allows other hardware constraints, such as the limited conductance states of non-volatile memory devices and limited fan-in connections in the neuromorphic architecture, to be incorporated easily during training. It hence greatly facilitates hardware-algorithm co-design and allows optimal performance to be achieved when deploying the trained SNN models onto the actual neuromorphic hardware. 

\begin{figure}[htb]	
	\centering
	\centerline
	{\includegraphics[width = 8 cm]{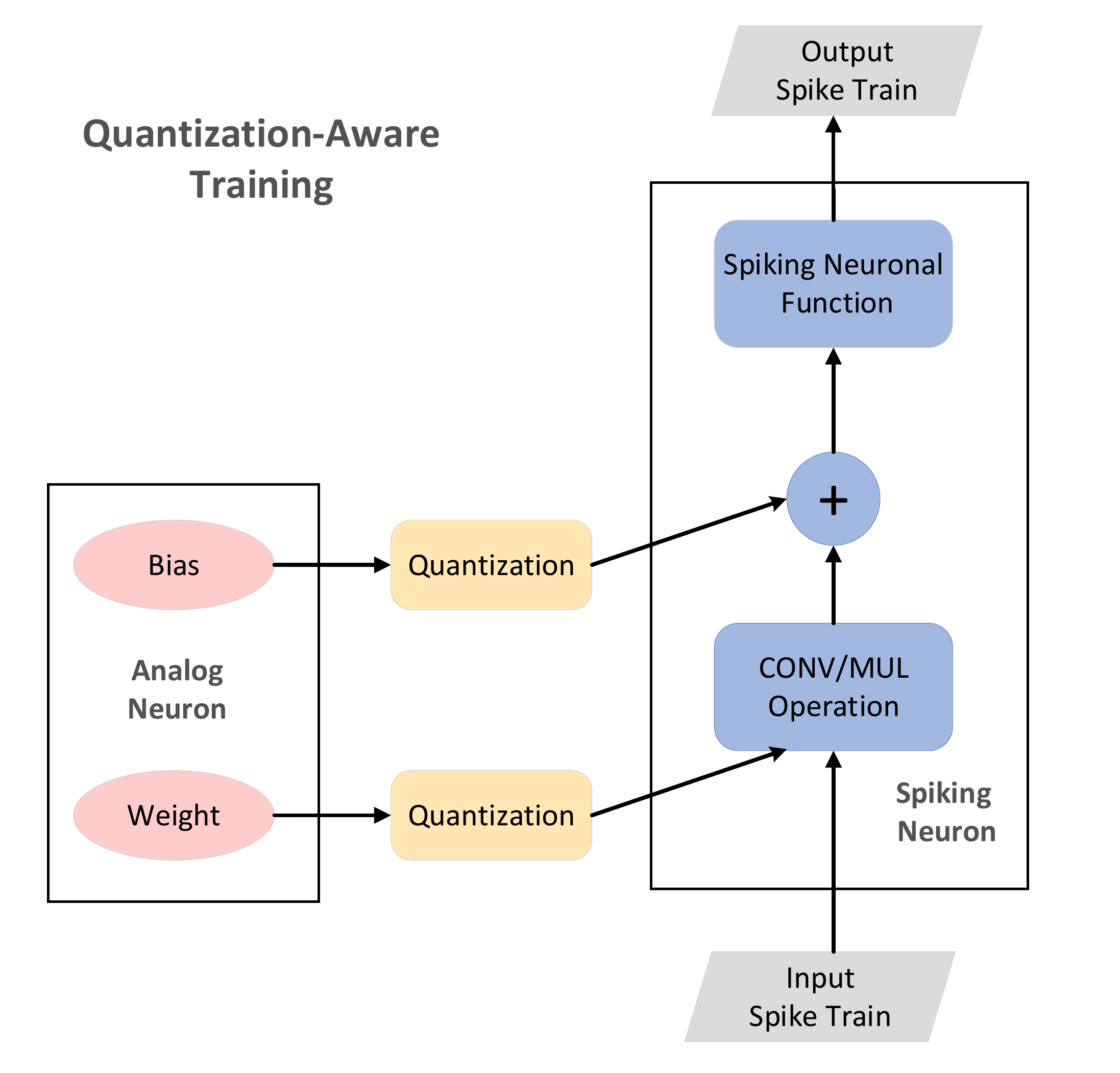}}
	\caption{Illustration of the quantization-aware training that can be incorporated into the proposed PTL framework. The full precision weight and bias terms of analog neurons are quantized to the desired precision before sharing with the coupled spiking neurons.}
	\label{fig:quantization_aware_training}
\end{figure}

To elucidate on this prospect, we explored the quantization-aware training \cite{jacob2018quantization} method whereby the low-precision weights are imposed progressively during training. As illustrated in Fig. \ref{fig:quantization_aware_training}, following the similar procedures that have been described for activation quantization in Eq. \ref{quantize}, the network weights and bias terms are quantized to a desirable precision before sharing to the SNN layer. While their full-precision copies are kept in the ANN layer to continue the learning with high precision. The flexibility provided by the PTL framework allows the SNN model to progressively navigate to a suitable parameter space to accommodate various hardware constraints. 

\section{Experiments on Pattern Classification}
\label{sec:experiment_classification}
In this section, we first investigate the scalability of spike-based learning methods, which motivates the proposal of a layer-wise learning method in fine-tuning the converted SNN. Secondly, we demonstrate the learning effectiveness and scalability of the proposed PTL framework on large-scale object recognition tasks. Thirdly, we investigate the effectiveness of the algorithm-hardware co-design methodology, that incorporates hardware constraints into the conversion process, with an example on the quantization-aware training for low precision neuromorphic hardware. Finally, we study the training efficiency of the proposed conversion framework as well as the improvements on the inference speed and energy efficiency of the trained SNN models. 

\subsection{Experimental Setup}
\label{exp_setup}
We perform all experiments with PyTorch library that supports accelerated and memory-efficient training on multi-GPU machines. Under a discrete-time simulation, we implement customized linear layer and convolution layer in Pytorch using IF neurons. We use the Adam optimizer \cite{kingma2014adam}  for all the experiments. To improve the training efficiency, we add batch normalization (BN) layer \cite{ioffe2015batch} after each convolution and linear layer. Following the approach introduced in \cite{ethImageNet}, we integrate the parameters of BN layers into their preceding convolution or linear layers' weights before sharing them with the coupled SNN layers. We use this setup consistently for both the pattern classification tasks of this section and the signal reconstruction tasks that will be presented in the next section unless otherwise stated. 

{\textbf{Dataset:}} We perform the object recognition experiments on the MNIST \cite{lecun1998gradient}, Cifar-10 \cite{krizhevsky2009learning} and ImageNet-12 datasets  \cite{deng2009imagenet}, which are widely used in machine learning and neuromorphic computing communities to benchmark different learning algorithms. The MNIST handwritten digits dataset consists of grayscaled digits of 28$\times$28 pixels that split into 60,000 training and 10,000 testing samples. The Cifar-10 dataset consists of 60,000 color images of size 32$\times$32$\times$3 from 10 classes, with a standard split of 50,000 and 10,000 for train and test, respectively. The large-scale ImageNet-12 dataset consists of over 1.2 million high-resolution images from 1,000 object categories. For Cifar-10 and MNIST datasets, we randomly split the original train set into train and validation sets with a split ratio of 90:10, which are fixed afterward for all the experiments. For ImageNet-12 dataset, the standard data split is followed for all experiments.

{\textbf{Network, Implementation and Evaluation Metric:}}
Two classical CNN architectures are explored on the Cifar-10 dataset: AlexNet \cite{krizhevsky2012imagenet} and VGG-11 \cite{simonyan2014very}. For the ImageNet-12 dataset, we performed experiments with AlexNet and VGG-16 \cite{simonyan2014very} architectures to facilitate comparison with other existing ANN-to-SNN conversion works. 

We also performed experiments with quantization-aware training of different weight precisions on the MNIST and Cifar-10 datasets. For MNIST dataset, the convolutional neural network with the structure of 28$\times$28-c16s1-c32s2-c32s1-c64s2-800-10 is used, wherein the numbers after `c' and `s' refer to the number of convolution filters and the stride of each convolution layer, respectively. The kernel size of 3 is used consistently for all convolution layers. For Cifar-10 dataset, we used AlexNet architecture.

For all experiments, the networks are trained for 100 epochs using the cross-entropy loss function. The patience period $T_p$ is adjusted based on the number of available training epoch and the network depth. The learning rate is initialized at $10^{-3}$ and decayed by 10 at Epoch 50. The classification accuracy is reported on the whole test set. 

\begin{figure*}[htb]	
	\centering
	\centerline
	{\includegraphics[width = 18 cm]{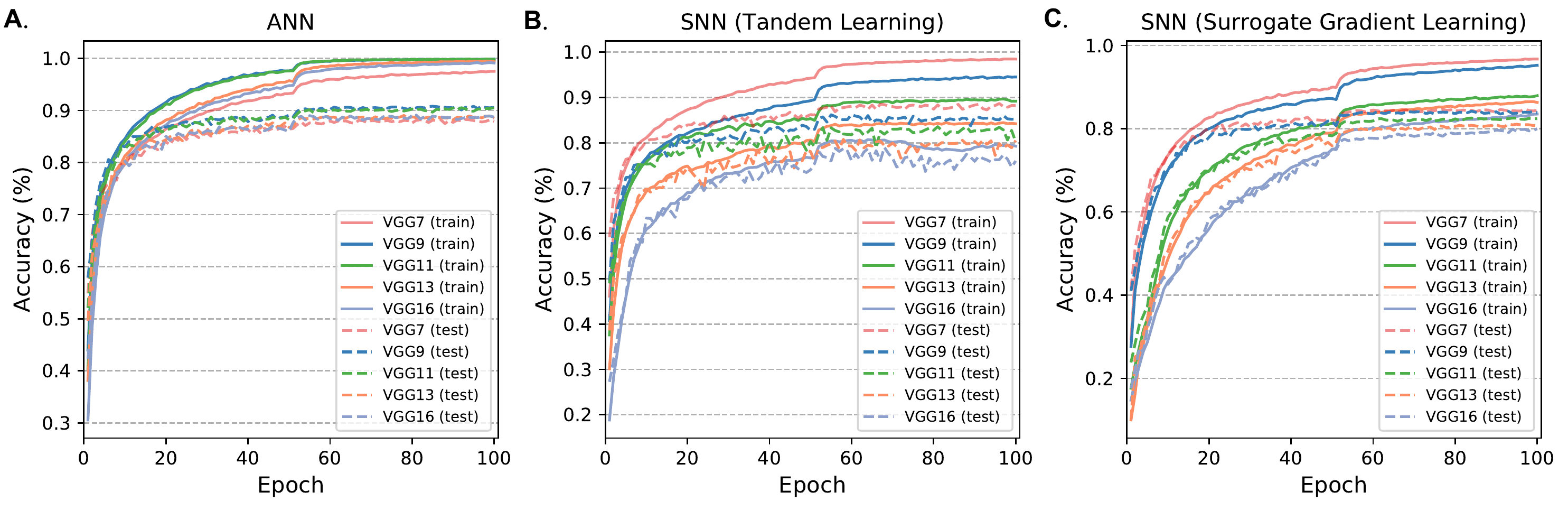}}
	\caption{Illustration of learning curves on the Cifar-10 dataset. (A) ANN models. (B) SNN models trained with spike count-based tandem learning~\cite{wu2019hybrid}. (C) SNN models trained with time-based surrogate gradient learning~\cite{wu2018direct}. It worth noting that the jump of learning curves at Epoch 50 is due to the learning rate decay.}
	\label{fig:learning_curve_compare}
\end{figure*}

\begin{table*}[!htb]
	\centering	
	\caption{Comparison of classification accuracy of different SNN implementations on the Cifar-10 and ImageNet-12 test sets. The numbers inside and outside the round bracket of the `Accuracy' column refer to the top-1 and top-5 accuracy, respectively.} 				
	\resizebox{16 cm}{!}{
		\begin{tabular}{llcccc}
			\toprule
			\toprule 
			&\textbf{Model} & \textbf{Network} &\textbf{Method} & \textbf{Accuracy (\%)} & \textbf{Time Steps}\\ 
			\toprule
			\parbox[t]{2mm}{\multirow{10}{*}{\rotatebox[origin=c]{90}{\textbf{Cifar-10}}}}
			&Wu et al. (2019) \cite{wu2018direct}  & AlexNet (SNN)& Surrogate Gradient Learning & 85.24 & -\\	
			&Hunsberger and Eliasmith (2016)\cite{hunsberger2016training} & AlexNet (SNN) & Constrain-then-Train & 83.54 & 200\\ 
			&\textbf{This work} & \textbf{AlexNet (ANN)} & \textbf{Error Back-propagation} &\textbf{89.59} & \textbf{16}\\					
			&\textbf{This work} & \textbf{AlexNet (SNN)} & \textbf{Progressive Tandem Learning} &\textbf{90.86} & \textbf{16}\\					
			&Rueckauer et al. (2017)\cite{ethImageNet} & VGG-like (SNN) & ANN-to-SNN conversion  & 88.82 & -\\													
			&Severa, William, et al. (2019)\cite{severa2019training}  & VGG-like (SNN) & Binary Neural Network & 84.67 & 1\\
			&Nitin et al. (2020)\cite{rathi2020enabling}  & VGG-16 (SNN) & ANN-to-SNN conversion & 90.20 & 100 \\	
			&Nitin et al. (2020)\cite{rathi2020enabling}  & VGG-16 (SNN) & ANN-to-SNN conversion + STDB  & 91.13 & 100 \\	
			&\textbf{This work} & \textbf{VGG-11 (ANN)} & \textbf{Error Back-propagation} &\textbf{90.59} & \textbf{16}\\					
			&\textbf{This work} & \textbf{VGG-11 (SNN)} & \textbf{Progressive Tandem Learning} &\textbf{91.24} & \textbf{16}\\
			\toprule 
			\toprule						
			\parbox[t]{2mm}{\multirow{10}{*}{\rotatebox[origin=c]{90}{\textbf{ImageNet}}}}
			&Hunsberger and Eliasmith (2016)\cite{hunsberger2016training} & AlexNet (SNN) & Constrain-then-Train  & 51.80 (76.20) & 200\\
			&Wu et al. (2019)\cite{wu2019hybrid} & AlexNet (SNN) & Tandem Learning  & 50.22 (73.60) & 10\\
			&\textbf{This work} & \textbf{AlexNet (ANN)} & \textbf{Error Back-propagation} &\textbf{58.53 (81.07)} & \textbf{16}\\														
			&\textbf{This work} & \textbf{AlexNet (SNN)} & \textbf{Progressive Tandem Learning} &\textbf{55.19 (78.41)} & \textbf{16}\\
			&Rueckauer et al. (2017)\cite{ethImageNet} & VGG-16 (SNN) & ANN-to-SNN conversion  & 49.61 (81.63) & 400  \\			
			&Sengupta et al. (2019)\cite{sengupta2019going}  & VGG-16 (SNN) & ANN-to-SNN conversion  & 69.96 (89.01) & 2500 \\
			&Nitin et al. (2020)\cite{rathi2020enabling}  & VGG-16 (SNN) & ANN-to-SNN conversion  & 68.12 (-) & 2500 \\
			&Nitin et al. (2020)\cite{rathi2020enabling}  & VGG-16 (SNN) & ANN-to-SNN conversion + STDB & 65.19 (-) & 250 \\
			&\textbf{This work} & \textbf{VGG-16 (ANN)} & \textbf{Error Back-propagation} &\textbf{71.65 (90.37)} & \textbf{16} \\				
			&\textbf{This work} & \textbf{VGG-16 (SNN)} & \textbf{Progressive Tandem Learning} &\textbf{65.08 (85.25)} & \textbf{16} \\		
			\toprule						
			\toprule
		\end{tabular}
		\label{results}
	}
\end{table*}

\subsection{Accumulated Errors with Spike-based Learning Methods}
\label{sec:spike_based_learning}
As discussed in Section \ref{sec:layer_wise_training}, to compensate for the errors arising from the primitive ANN-to-SNN conversion, a training method is required to fine-tune the network weights. 
Here, we take the object recognition task on the Cifar-10 dataset as an example to study the scalability of spike-based learning methods in training deep SNNs to perform rapid pattern recognition. Specifically, we implemented the surrogate gradient learning method and tandem learning method proposed in \cite{neftci2019surrogate} and \cite{wu2019hybrid}, respectively. The network structures employed in this study are taken from the famous VGGNet \cite{simonyan2014very}.

With an encoding time window $N_s$ of 8, the learning curves for ANN and SNN models with different network depth are presented in Fig. \ref{fig:learning_curve_compare}. As shown in Fig. \ref{fig:learning_curve_compare}A, the training converges easily for all ANN models, despite slight overfitting observed for the VGG13 and VGG16 models. In contrast, the training convergence is difficult for the spiking counterparts that have a network depth of over 10 layers as shown in Figs. \ref{fig:learning_curve_compare}B and \ref{fig:learning_curve_compare}C. This observation suggests the gradient approximation error tends to accumulate over layers with the spike-based learning methods and significantly degrades the learning performance for deep SNNs over 10 layers. In the following sections, we will show that the proposed PTL framework that performs fine-tuning one layer at a time can effectively overcome the accumulated gradient approximation errors and scale-up freely to deep SNNs with 16 layers.

\begin{figure*}[htb]	
	\centering
	\centerline
	{\includegraphics[width = 18 cm]{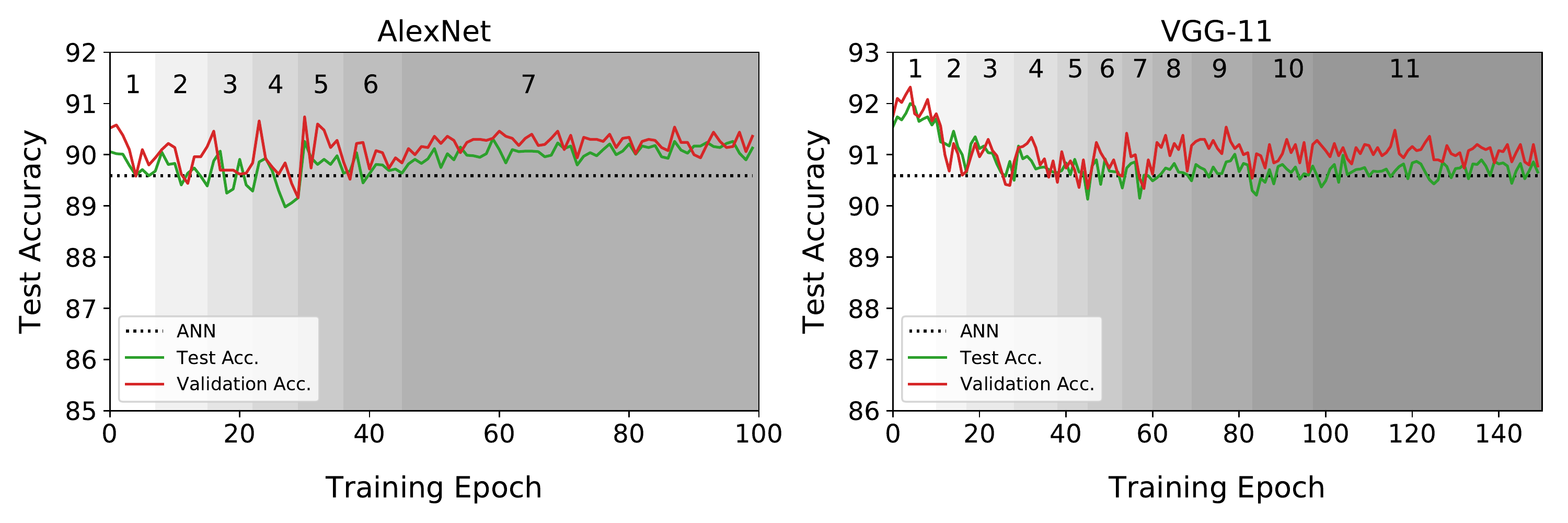}}
	\caption{Illustration of the training progresses of the AlexNet and VGG-11 on the Cifar-10 dataset ($N_s=16$, $T_p=6$). The shaded regions correspond to different training stages. After replacing each ANN layer with an equivalent SNN layer at the beginning of each training stage, the validation and test accuracies can be quickly restored with the proposed PTL framework.}
	\label{learning_progress}
\end{figure*}

\subsection{Object Recognition on Cifar-10 and ImageNet-12}
As shown in Fig. \ref{learning_progress}, we plot the training progress of the AlexNet and VGG-11 models on the Cifar-10 dataset, to illustrate the effectiveness of the proposed PTL framework. As expected, the validation accuracy drops mostly at the beginning of each conversion stage due to the conversion errors introduced. Notably, these errors are counteracted by the proposed layer-wise learning method, whereby the test and validation accuracies restored quickly with only a few training epochs. Overall, the validation and test accuracies remain relatively stable during the whole training progress and surpass those of the pre-trained ANNs after training. It suggests that the proposed conversion framework can significantly reduce the representation space $N_s$ by exploiting the redundancies existed in the high-dimensional feature representation of the ANN.

As reported in Table \ref{results}, the trained deep SNNs achieve state-of-the-art classification accuracies over other existing SNN implementations with similar network architecture, with a test accuracy of 90.86\% and 91.24\% for AlexNet and VGG-11 respectively on the Cifar-10 dataset. It is worth mentioning that these SNN models even outperform their pre-trained ANN baselines by 1.27\% and 0.65\%. In comparison with a recently introduced binary neural network training method for neuromorphic implementation \cite{severa2019training}, which achieved a classification accuracy of 84.67\%, the results suggest that the larger encoding time window $N_s=16$ contributes to the higher accuracy.

To study the scalability of the proposed PTL framework on more complex datasets and network architectures, we conduct experiments on the challenging ImageNet-12 dataset. Due to the high computational complexity of modeling deep SNNs and the huge memory demand to store their intermediate states, only a limited number of ANN-to-SNN conversion methods have achieved some promising results on this dataset. 

As in Table \ref{results}, the spiking AlexNet and VGG-16 models trained with the proposed PTL framework achieve promising results on the ImageNet-12 dataset. For the spiking AlexNet, the top-1 (top-5) accuracy improves by 3.39\% (2.21\%) over the early work that takes a constrain-then-train approach \cite{hunsberger2016training}. Meanwhile, the total number of time steps required is reduced by more than one order from 200 to 16. For the spiking VGG-16, despite the total number of time steps reduced by at least 25 times, our result is as competitive as those achieved with the state-of-the-art ANN-to-SNN conversion approaches \cite{ethImageNet,sengupta2019going}. 

Nitin et al. \cite{rathi2020enabling} recently apply a spike-based learning method to fine-tune the weights of the converted SNN end-to-end, so as to speed up the model at run time. This method successfully reduces the total time steps from 2,500 to 250, with accuracy drops by about 3\% on the ImageNet-12 dataset. In contrast, the discrete neural representation proposed in this work provides an improved network initialization that allows for a more radical reduction in the encoding time window. Notably, the classification accuracy of our system is on par with theirs, while requiring only a total of 16 time steps. Although our results drop from the pre-trained AlexNet and VGG-16 models by about 3\% and 6\% respectively, it is expected that this gap could be closed by increasing the encoding time window $N_s$.   

\subsection{Quantization-Aware Training for Low Precision Neuromorphic Hardware}
Table \ref{tab:quantization_aware} provides the object recognition results with the quantization-aware training. On the MNIST and Cifar-10 datasets, the low-precision SNN models perform exceedingly well regardless of the reduced bit-width and the limited representation space (i.e., $N_s=16$). Specifically, when the weights are quantized to 4-bit, the classification accuracy drops  by only 0.03\% and 0.85\% on the MNIST and Cifar-10 datasets, respectively. Therefore, the proposed PTL framework offers immense opportunities for implementing SNNs on the low-precision neuromorphic hardware, for instance with emerging non-volatile memory devices that suffering from limited conductance states.

\begin{table}[htb]
	\centering
	\caption{Comparison of the classification results as a function of weight precision. The result of SNN models is obtained through quantization-aware training. The average results across 5 independent runs are reported.}
	\vspace{5 mm}
	\resizebox{9 cm}{!}{
		\begin{tabular}{| c c | c c c|}
			\hline
			\multicolumn{2}{|c|}{\bf{Benchmark}} & \bf{Bit Width} & \bf{Acc. (\%)} & \bf{Change of Acc. (\%)}\\
			\hline
			\hline
			\multicolumn{2}{|c|}{\multirow{6}{*}{\bf{MNIST}}} & Float32 & 99.32 & 0 \\ 
			\multicolumn{2}{|c|}{}  & 8-bit & 99.32 & 0 \\
			\multicolumn{2}{|c|}{}  & 7-bit & 99.30 & -0.02 \\
			\multicolumn{2}{|c|}{}  & 6-bit & 99.29 & -0.03\\
			\multicolumn{2}{|c|}{}  & 5-bit & 99.30 & -0.02\\
			\multicolumn{2}{|c|}{}  & 4-bit & 99.29 & -0.03\\			
			\hline
			\hline
			\multicolumn{2}{|c|}{\multirow{6}{*}{\bf{Cifar-10}}} & Float32 & 90.33  & 0 \\ 
			\multicolumn{2}{|c|}{} & 8-bit & 90.11  & -0.22 \\
			\multicolumn{2}{|c|}{} & 7-bit & 90.06 & -0.27 \\
			\multicolumn{2}{|c|}{} & 6-bit &  90.07 & -0.26 \\		
			\multicolumn{2}{|c|}{}  & 5-bit & 90.04 & -0.29 \\
			\multicolumn{2}{|c|}{}  & 4-bit & 89.48 & -0.85 \\
			\hline
		\end{tabular}
	}
	\label{tab:quantization_aware}
\end{table}

\begin{figure*}[htb]	
	\centering
	\centerline
	{\includegraphics[width = 18 cm]{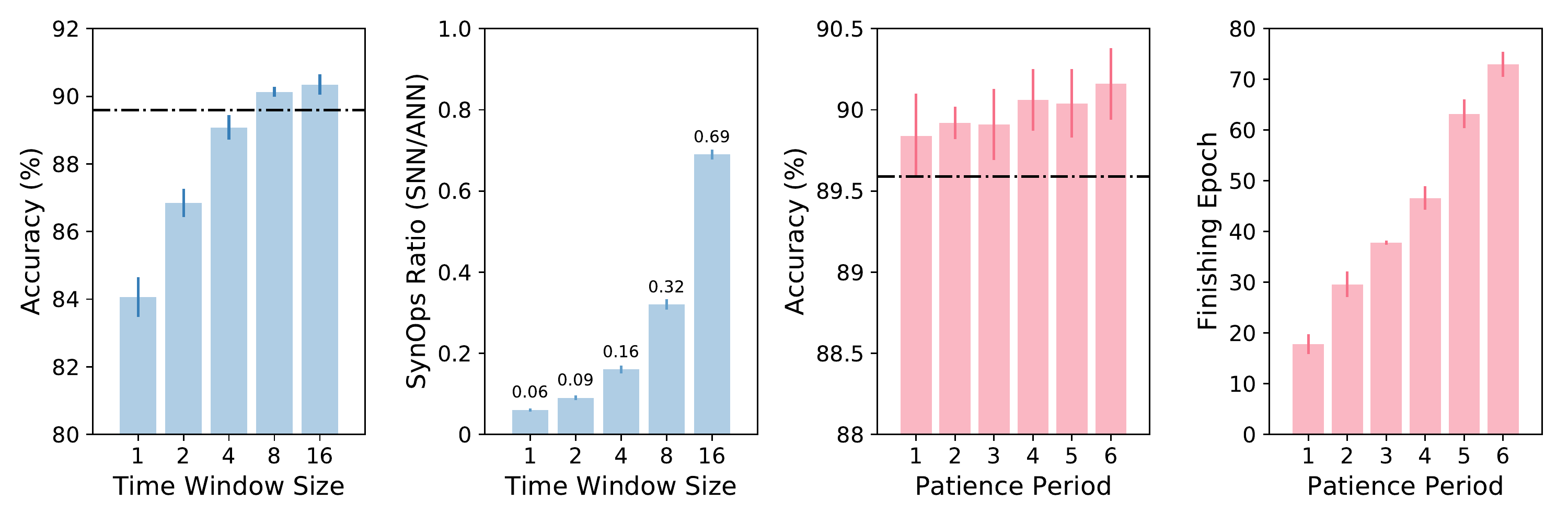}}
	\caption{(A) Classification accuracy as a function of the encoding time window on the Cifar-10 dataset. The horizontal dashed line refers to the accuracy of the pre-trained ANN. (B) The ratio of total synaptic operations between SNN and ANN as a function of encoding time window on the Cifar-10 dataset. (C)  Classification accuracy as a function of the patience period defined in the adaptive scheduler. (D) Finishing epoch as a function of the patience period. All  experimental results are summarized over 5 independent runs with spiking AlexNet.}
	\label{Tencode_Energy_Analysis}
\end{figure*}

\subsection{Rapid and Efficient Classification with SNN}
When implemented on the neuromorphic chips, the SNNs have great potential to improve the real-time performance and energy efficiency over ANNs. However, the learning methods grounded on the firing rate assumption require long inference time, typically a few hundred to thousands of time steps, to reach a stable network firing state. They diminish the latency advantages that can be obtained from the asynchronous operation of SNNs. In contrast, the proposed conversion framework allows making efficient use of the available time steps, such that rapid inference can be performed with only 16 time steps on the ImageNet-12 dataset. As shown in Fig. \ref{Tencode_Energy_Analysis}(A), we notice a clear positive correlation between the encoding time window size and the classification accuracy on the Cifar-10 dataset. Notably, a reliable prediction can still be made with only a single time step when SNN is trained to utilize this limited amount of information as in the scenario of binary neural networks, while the performance can be further improved when larger encoding time windows are provided.

To further study the energy efficiency of trained SNN models, we follow the convention by counting the synaptic operations per inference and calculating the ratio to the corresponding ANN models \cite{wu2019hybrid,ethImageNet}. In general, the total synaptic operations required by the ANN is a constant number depending on the network architecture, while it positively correlates with the encoding time window and the firing rate for SNNs. As shown in  Fig. \ref{Tencode_Energy_Analysis}(B), under the iso-accuracy setting, when the ANN and SNN models achieve equal accuracy, the SNN ($N_s = 8$) consumes only around 0.315 times total synaptic operations over ANN. In contrast, the state-of-the-art
SNN implementations with the ANN-to-SNN conversion
and spike-based learning methods have reported a SynOps ratio of 25.60 and 3.61 respectively on a similar VGGNet-9 network \cite{lee2019enabling}. It suggests our SNN implementation is  81.27 and 11.46 times more efficient at run-time respectively. 

It is worth noting that SNNs perform mostly accumulate (AC) operations to integrate the membrane potential contributions from incoming spikes. In contrast, multiply-accumulate (MAC) operations are used in ANN which is significantly more expensive in terms of energy consumption and chip area usage. For instance, the simulations in a Global Foundry 28 nm process report the MAC operation is 14x costly than the AC operation and requires 21x chip area \cite{ethImageNet}. Therefore, over 40 times cost savings can be received from SNN models by taking the sparse and cheap AC operations over the ANN counterparts, and the cost savings can be further boosted from efficient neuromorphic chip architecture design and emerging ultra-low-power devices implementation.

Figs. \ref{Tencode_Energy_Analysis}(C) and \ref{Tencode_Energy_Analysis}(D) present the classification results and the required training epochs as a function of the patience period in the adaptive training scheduler. As shown in Fig.  \ref{Tencode_Energy_Analysis}(C), a competitive classification accuracy that surpasses the pre-trained ANN model can be achieved even with a patience period of only 1, that requires an average epoch of only 18 as shown in Fig. \ref{Tencode_Energy_Analysis}(D). The accuracy can be further improved if a longer patience period is given. 

\begin{figure*}[htb]	
	\centering
	\centerline
	{\includegraphics[width = 16 cm]{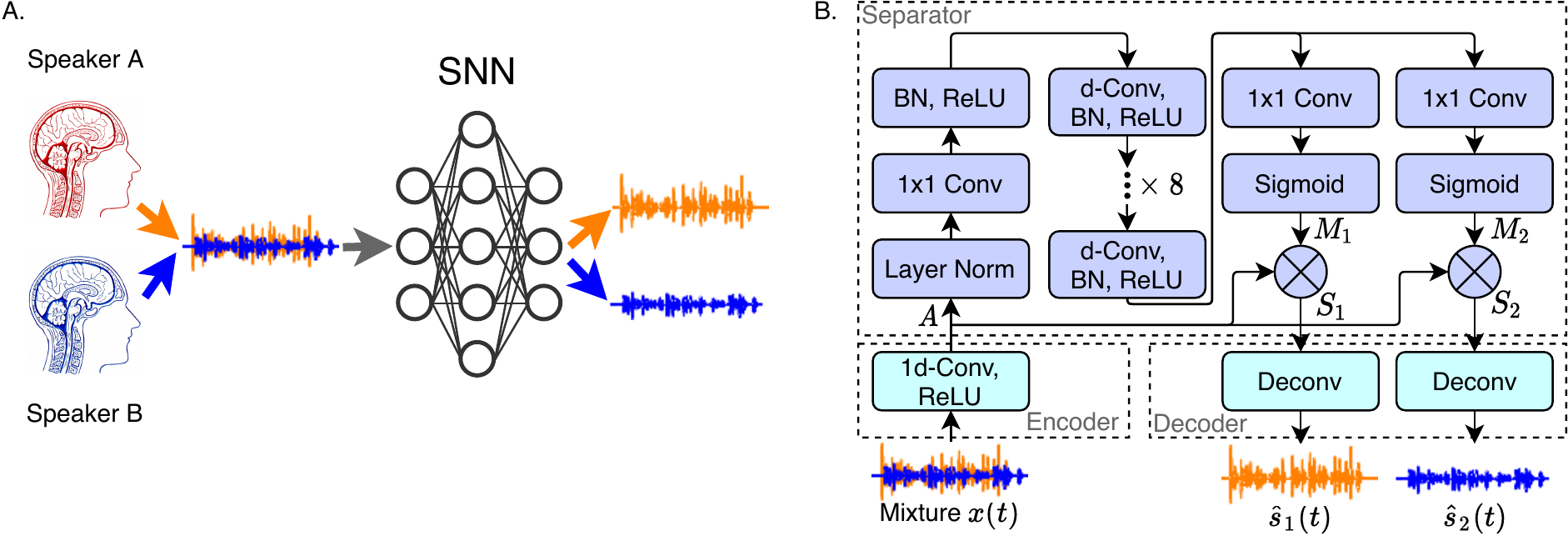}}
	\caption{(A) Illustration of the SNN-based speech separation approach to solving the cocktail party problem. (B) Illustration of the proposed SNN-based speech separation network. It takes two speakers mixture as input and outputs two independent streams for each individual speaker. ``1d-Conv" indicates a 1-dimensional convolution. ``1$\times$1 Conv" is a convolution with a 1$\times$1 kernel. ``d-Conv" is a dilated convolution. ``Deconv" is a deconvolution (also known as transposed convolution). ``ReLU" is a rectified linear unit function. ``BN" represents batch normalization. $\otimes$ refers to the element-wise multiplication.}
	\label{fig:speech_seperation_network}
\end{figure*}

\section{Experiments on Signal Reconstruction}
\label{sec:experiment_regression}
In Section \ref{sec:experiment_classification}, we demonstrate superior learning capability and scalability of the proposed PTL framework on pattern classification tasks. The existing ANN-to-SNN conversion works mainly focus on the pattern classification tasks, where a high-precision output is not required. The regression tasks like signal reconstruction however require the SNN model to predict high precision outputs using spikes, that have not been well explored. In this section, we further apply SNNs to solve pattern regression tasks that are known to be challenging for SNNs. Specifically, we perform experiments on the image reconstruction and speech separation tasks, both of which require to reconstruct high-fidelity signals.

\subsection{Image Reconstruction with Autoencoder}
An autoencoder is a type of neural network that learns to decompose input signals into a compact latent representation, and then use that representation to reconstruct the original signals as closely as possible~\cite{Goodfellow-et-al-2016}. Typically an autoencoder learns the compact latent representation through a bottleneck layer that has a reduced dimensionality over the input. In this way, it ignores the variation, removes the noise, and disentangles a mixture of information. Here, we investigate the compact latent representation extraction and  reconstruction for static images using spike counts.

\subsection{Time-domain Speech Separation}
\label{sec:speech_separation}
Speech separation is one of the solutions for the cocktail party problem, where one is expected to selectively listen to a particular speaker in a multi-talker scenario~\cite{selectiveatt}.  Physiological studies reveal that selective auditory attention takes place both locally by transforming the receptive field properties of individual neurons and globally throughout the auditory cortex by rapid neural adaptation, or plasticity, of the cortical circuits~\cite{isomura2015cultured, selectiveatt2}. However, machines have yet to achieve the same attention ability as humans in segregating mixed stimuli into different streams. Such auditory attention capability is highly demanded in real-world applications, such as, hearing aids \cite{wang2017deep}, speech recognition \cite{li2015robust}, speaker verification\cite{rao2019target}, and speaker diarization\cite{sell2018diarization}.

Inspired by the recent progress in deep ANN approaches to time-domain speech separation and extraction \cite{8707065, spex}, we propose and implement a deep SNN-based solution for speech separation.  As shown in Fig. \ref{fig:speech_seperation_network}, the SNN takes the mixture speech as input and generates individual speech into separate streams. With a stack of dilated convolutional layers, the SNN captures the long-range dependency of the speech signal with a manageable number of parameters. It is optimized to maximize a scale-invariant signal-to-distortion ratio (SI-SDR) \cite{le2019sdr} loss for high fidelity speech reconstruction.

The proposed SNN-based speech separation framework consists of three components: an encoder, a separator, and a decoder, as shown in Fig. \ref{fig:speech_seperation_network}. The encoder transforms the time-domain mixture signal into a high-dimensional representation, which is then taken as the input to the separator. The separator estimates a mask for each speaker at each time step. After that, a suitable representation for every individual speaker is extracted by filtering the encoded representation of the input mixture with the estimated mask for that speaker. Finally, the time-domain signal of each speaker is reconstructed using a decoder.

\subsection{Experimental Setup}
In the following, we will present the experiments designed for image reconstruction and speech separation tasks. By applying the PTL framework, the pre-trained ANNs are converted into SNNs for high-fidelity signal reconstruction in these tasks.

\subsubsection{Image Reconstruction}
\paragraph{\textbf{Dataset}} The MNIST dataset \cite{lecun1998gradient} is used for the image reconstruction task, which consists of 60,000 training and 10,000 test samples. These samples are directly used for training and testing without applying any data pre-processing steps.

\paragraph{\textbf{Network, Implementation and Evaluation Metric}}
We evaluate a fully-connected autoencoder that has an architecture of 784-128-64-32-64-128-784, wherein the numbers refer to the number of neurons at each layer \cite{severa2019training}. The sigmoid activation function is used in the output layer to normalize the output so as to match to the input range, while the rest of the layers use a ReLU activation function. Following the neural coding scheme introduced in Section \ref{subsec:neural_code}, instead of using the spike count, the free aggregate membrane potential of spiking neurons in the final SNN layer is considered as the pre-activation quantity to the sigmoid activation function, which provides a high-resolution reconstruction. 

The networks are trained for 100 epochs using the mean square error (MSE) loss function, and the patience period $T_p$ of the training scheduler is set to 6. We report the MSE of reconstructed images on the MNIST test set with different encoding time window size. The rest of the training configurations follow those used in pattern classification tasks as presented in Section \ref{exp_setup}. 

\subsubsection{Time-domain Speech Separation}
\paragraph{\textbf{Dataset}}
We evaluated the methods on the two-talker mixed WSJ0-2mix dataset\footnote{Available at: http://www.merl.com/demos/deep-clustering. The database used in this work is simulated with the released script and configuration in \cite{hershey2016deep}.} \cite{hershey2016deep} with a sampling rate of 8kHz, which was mixed by randomly choosing utterances of two speakers from the WSJ0 corpus \cite{garofolo1993csr}. The WSJ0-2mix corpus consists of three sets: training set ($20,000$ utterances $\approx$ $30h$), development set ($5,000$ utterances $\approx$ $8h$), and test set ($3,000$ utterances $\approx$ $5h$). Specifically, the utterances from $50$ male and $51$ female speakers in the WSJ0 training set (si\_tr\_s) were randomly selected to generate the training and development set in WSJ0-2mix at various signal-to-noise (SNR) ratios that uniformly chosen between 0dB and 5dB. Similarly, the test set was created by randomly mixing the utterances from $10$ male and $8$ female speakers in the WSJ0 development set (si\_dt\_05) and evaluation set (si\_et\_05). The test set was considered as the open condition evaluation because the speakers in the test set were different from those in the training and development sets. We used the development set to tune parameters and considered it as the closed condition evaluation because the speakers are seen during training. The utterances in the training and development set were broken into $4$s segments.

\paragraph{\textbf{Network and Implementation}} Inspired by the Conv-TasNet speech separation system \cite{8707065}, the proposed SNN-based speech separation system first encodes the mixture input $x(t)\in\mathbb{R}^{1\times T}$ by a 1d-convolution with $N(=512)$ filters followed by the ReLU activation function. Each filter has a window of $L(=20)$ samples with a stride of $L/2(=10)$ samples. In the separator part, a mean and variance normalization with trainable gain and bias parameters is applied to the encoded representations $A\in\mathbb{R}^{K\times N}$ on the channel dimension, where $K$ is equal to $2(T-L)/L+1$. A 1$\times$1 convolution together with batch normalization and ReLU activation is applied to the normalized encoded representations. The dilated convolutions with $512$ filters are repeated $10$ times with dilations ratios of $[2^0, 2^1, ..., 2^9]$. These dilated convolution filters have a kernel size of $1\times 3$ and a stride of $1$. The batch normalization and ReLU activation function are also applied to the dilated convolutions layers. A mask ($M_1$, $M_2$) for each speaker is then estimated by a 1$\times$1 convolution with a sigmoid activation function. The modulated representation ($S_1$,$S_2$) for each speaker is obtained by filtering the encoded representation $A$ with the estimated mask ($M_1$, $M_2$). Finally, the time-domain signal ($s_1$,$s_2$) for each speaker is reconstructed by the decoder, which acts as the inverse process of the encoder.

The ANN-based system is optimized with the learning rate started from $0.001$ and is halved when the loss increased on the development set for at least $3$ epochs. Then, we take the pre-trained ANN model and convert the separator into an SNN. It is worth mentioning that the aggregate membrane potential is applied as the inputs to the last 1$\times$1 convolution layer where a float-point representation is required to generate high-resolution auditory masks. The encoding time window $N_s$ and patience period $T_p$ are set to 32 and 3 for SNNs, respectively. Both ANN and SNN models are trained for 100 epochs, and an early stopping scheme is applied when the loss does not improve on the development set for $10$ epochs. 

\paragraph{\textbf{Training Objective and Evaluation Metric}} The speech separation system is optimized by maximizing the scale-invariant signal-to-distortion ratio (SI-SDR) \cite{le2019sdr}, that is defined as:
\begin{equation} 
\label{si-sdr}
\text{SI-SDR} = 10\log_{10}\left(\frac{||\frac{\langle\hat{s}, s\rangle}{\langle s,s\rangle}s||^2}{||\frac{\langle\hat{s}, s\rangle}{\langle s,s\rangle}s-\hat{s}||^2}\right)
\end{equation}
where $\hat{s}$ and $s$ are separated and target clean signals, respectively.  $\langle\cdot,\cdot\rangle$ denotes the inner product. To ensure scale invariance, the signals $\hat{s}$ and $s$ are normalized to zero-mean prior to the SI-SDR calculation. Since we don't know which speaker the separated stream belongs to (permutation problem), we adopt permutation invariant training to find the best permutation by maximizing the SI-SDR performance among all the permutations. The SI-SDR is used as the evaluation metric to compare the performances of the original ANN-based and the converted SNN-based speech separation systems. We also evaluate the systems with Perceptual Evaluation of Speech Quality (PESQ) \cite{rix2001perceptual,hu2007evaluation}, which is recommended as the ITU-T P.862 standard to automatically assess the speech quality instead of the subjective Mean Opinion Score (MOS). During the evaluation, the permutation problem between the separated streams and the corresponding target clean signals are decided following the permutation invariant training during the training phase. 

\subsection{Experimental Results}

\subsubsection{Image Reconstruction with Autoencoder}
\begin{figure}	
	\centering
	\centerline
	{\includegraphics[width = 8.5 cm]{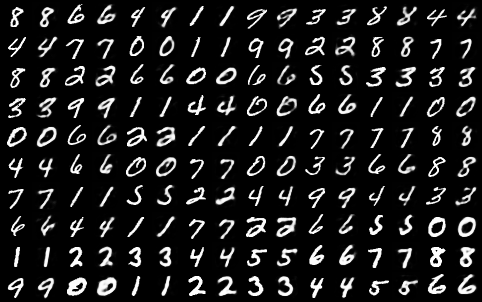}}
	\caption{Illustration of the reconstructed images from spiking autoencoder ($N_s=32$) on the MNIST dataset. For each pair of digits, the left side is the original image and the right side is the reconstruction by SNN.}
	\label{ae_result}
\end{figure}

Table \ref{tab:image_construction} provides the image reconstructions results. As expected, a clear negative correlation between the encoding time window size $N_s$ and the MSE has been observed. Notably, with an encoding time window of 32, the spiking autoencoder achieves an MSE of 0.00662 on the MNIST dataset, which is a slight improvement from 0.00667 of the pre-trained ANN. As also shown in  Fig. \ref{ae_result}, this spiking autoencoder ($N_s=32$)  can effectively reconstruct images with high quality. In contrast to the object recognition results shown in Fig. \ref{Tencode_Energy_Analysis}(A), the results on the image reconstruction suggest regression tasks may require a larger discrete representation space or encoding time window to match the performance of the pre-trained ANN. 

\begin{table}[htb]
	\centering
	\caption{Comparison of the image reconstruction results as a function of the encoding time window size $N_s$. The average results across 5 independent runs are reported.}
	\resizebox{8.5 cm}{!}{
		\begin{tabular}{| c c | c c c|}
			\hline
			\multicolumn{2}{|c|}{\bf{Model}} & $\bf{N_s}$ & \bf{MSE} & \bf{Change of MSE}\\
			\hline
			\hline
			\multicolumn{2}{|c|}{\bf{ANN}}  & - & 0.00667 & - \\
			\hline
			\hline
			\multicolumn{2}{|c|}{\multirow{6}{*}{\bf{SNN}}} & 32 & 0.00662  & -0.00005 \\ 
			\multicolumn{2}{|c|}{}  & 16 & 0.01720 & 0.01053 \\
			\multicolumn{2}{|c|}{}  & 8 & 0.02361 & 0.01694 \\
			\multicolumn{2}{|c|}{}  & 4 & 0.02724 & 0.02057 \\
			\multicolumn{2}{|c|}{}  & 2 & 0.03435 & 0.02768 \\
			\multicolumn{2}{|c|}{}  & 1 & 0.04032 & 0.03365 \\
			\hline			
		\end{tabular}
	}
	\label{tab:image_construction}
\end{table}

\begin{figure}[htb]	
	\centering
	\centerline
	{\includegraphics[width = 9 cm]{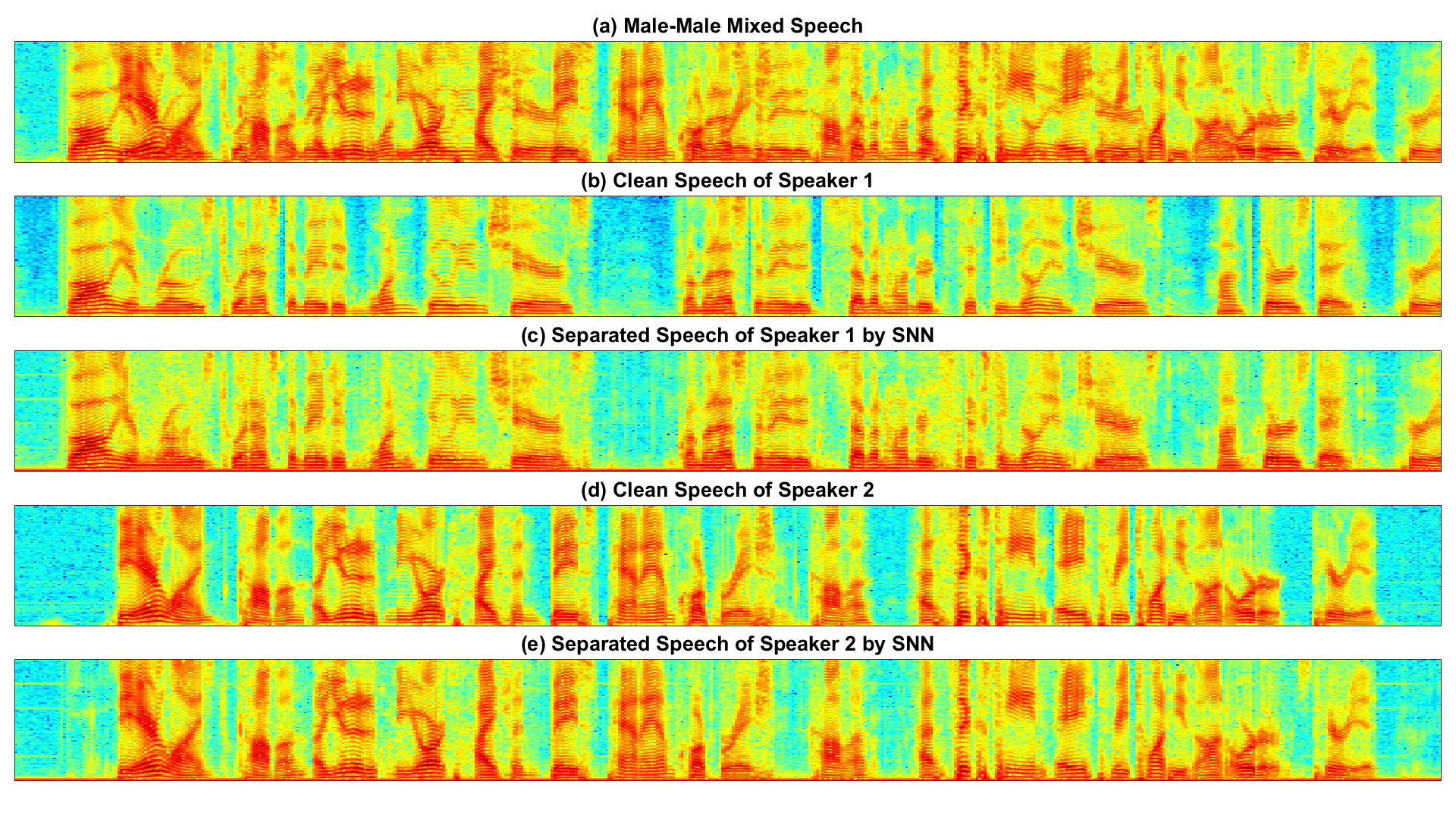}}
	\caption{The example of male-male mixture speech separated by SNN-based speech separation network.}
	\label{fig:seperation_male_male}
\end{figure}


\subsubsection{Time-domain Speech Separation}
Table \ref{tab:speech_separation} summarizes the comparative study between the original ANN-based and the converted SNN-based speech separation systems. The ANN- and SNN-based systems achieve an SI-SDR of $12.8$ dB and $12.2$ dB under the open condition evaluation, respectively. In terms of the perceptual quality, we observe that the ANN and SNN have a very close PESQ score of $2.94$ and $2.85$, respectively. The open condition evaluation results suggest that the SNN can achieve comparable performance to the ANN in this challenging speech separation task, while the SNN can take additional benefits of rapid inference and energy efficiency at test time. The same conclusion could also be drawn for the closed condition evaluation.

\begin{table}[tb]
	\centering
	\caption{Comparative study  between ANN and SNN on speech separation tasks under both  closed and open condition. The closed condition is on the development set, where the speakers are seen during training. The open condition is on the test set, where the speakers are unseen during training. ``Diff." refers to the different gender mixture. ``Same" refers to the same gender mixture. ``Overall" refers to the combination of both different and same gender mixtures.}
	\resizebox{9 cm}{!}{
		\begin{tabular}{|c| c | c c c| c c c|}
		\hline
		\multirow{2}{*}{\bf{Cond.}} & \multirow{2}{*}{\bf{Methods}} & \multicolumn{3}{c|}{\bf{SI-SDR (dB)}} & \multicolumn{3}{c|}{\bf{PESQ}} \\ \cline{3-8}
		 & & \bf{Diff.} & \bf{Same} & \bf{Overall} & \bf{Diff.} & \bf{Same} & \bf{Overall} \\ \hline \hline
		\multirow{2}{*}{\bf{Closed}} & ANN & 15.2 & 11.7 & 13.5 & 3.12 & 2.83 & 2.97 \\
		 & SNN & 14.5 & 11.0 & 12.8 & 3.03 & 2.75 & 2.89 \\ \hline
		 \multirow{2}{*}{\bf{Open}} & ANN & 14.9 & 10.4 & 12.8 & 3.11 & 2.74 & 2.94 \\
		 & SNN & 14.2 & 9.8 & 12.2 & 3.02 & 2.66 & 2.85 \\ \hline
		\end{tabular}
	}
	\label{tab:speech_separation}
\end{table}

By listening to the separated examples generated by both ANN and SNN, we observe that the separated examples by SNN are very similar to those generated by ANN with high-fidelity. We publish some examples from the testing set (open condition) online to demonstrate our system performance \footnote{The listening examples are available at https://xuchenglin28.github.io/files/iccbc2019/index.html}. We randomly select a speech sample under the male-male mixture condition from the test set and show their magnitude spectra in Fig. \ref{fig:seperation_male_male}. We observe that the SNN obtains a similar spectrum as the ground truth clean spectrum even under the challenging condition of the same gender, where the multi-talkers have similar acoustic characteristics, i.e., pitch, hence less information is available to discriminate them from each other.

\section{Conclusion}
\label{sec:conclusion}
In this work, we reinvestigate the conventional ANN-to-SNN conversion approach and identify the accuracy and latency trade-off with the adopted firing rate assumption. Taking inspiration from the activation quantization works, we further propose a novel network conversion method, whereby spike count is utilized to represent the activation space of analog neurons. This configuration allows better exploitation of the limited representation space and improves the inference speed. Furthermore, we introduce a layer-wise learning method to counteract the errors resulted from the primitive network conversion. The proposed conversion and learning framework, that is called \textit{progressive tandem learning} (PTL), is highly automated with the proposed adaptive training scheduler, which supports flexible and efficient training. Benefiting from the proposed PTL framework, the algorithm-hardware co-design can be effectively accomplished by imposing the hardware constraints progressively during training. 

The SNNs thus trained have demonstrated competitive classification and regression capabilities on the challenging ImageNet-12 object recognition, image reconstruction, and speech separation tasks. Moreover, the proposed PTL framework allows making efficient use of the available encoding time window, such that rapid and efficient pattern recognition can be achieved with deep SNNs. Taking the quantization-aware training as an example, we illustrate how the hardware constraint, limited weight precision, can be effectively introduced during training, such that the optimal performance can be achieved on the actual neuromorphic hardware. By integrating the algorithmic power of deep SNNs and energy-efficient neuromorphic computing architecture, it opens up a myriad of opportunities for rapid and efficient inference on the pervasive low-power devices.

\ifCLASSOPTIONcompsoc
  \section*{Acknowledgments}
\else
  \section*{Acknowledgment}
\fi

J.~Wu and H.~Li are supported by the National Research Foundation, Singapore under its AI Singapore Programme (Award No: AISG-GC-2019-002) and (Award No: AISG-100E-2018-006), and its National Robotics Programme (Grant No. 192 25 00054), and by RIE2020 Advanced Manufacturing and Engineering Programmatic Grants A1687b003, and A18A2b0046. J.~Wu is also partially supported by the Zhejiang Lab’s International Talent Fund for Young Professionals.

\bibliographystyle{IEEEtran}
\bibliography{citation-main}

\ifCLASSOPTIONcaptionsoff
  \newpage
\fi

\end{document}